\documentclass{article}

\usepackage[english]{babel}

\usepackage[letterpaper,top=2cm,bottom=2cm,left=3cm,right=3cm,marginparwidth=1.75cm]{geometry}

\usepackage{amsmath}
\usepackage{graphicx}
\usepackage[colorlinks=true, allcolors=blue]{hyperref}

\title{Automated Classification of Dry Bean Varieties Using XGBoost and SVM Models}
\author{Ramtin Ardeshirifar\\
Department of Informatics,\\
University of Sussex\\
\href{mailto:R.ardeshirifar@sussex.ac.uk}{r.ardeshirifar@sussex.ac.uk}}
\date{}

\usepackage{titlesec}

\titleformat{\paragraph}
{\normalfont\normalsize\bfseries}{\theparagraph}{1em}{}
\titlespacing*{\paragraph}
{0pt}{3.25ex plus 1ex minus .2ex}{1.5ex plus .2ex}

\begin{document}
\maketitle

\section{Introduction}

This report focuses on the development of two multiclass classification models using computer vision and machine learning techniques to classify dry beans. The study aims to provide a method for obtaining uniform seed varieties from crop production, which is in the form of population, so the seeds are not certified as a sole variety. The primary objective is to distinguish seven different registered varieties of dry beans with similar features in order to obtain uniform seed classification.

The dataset used in this study consists of images of 13,611 grains of seven different registered dry beans, which were subjected to segmentation and feature extraction stages, resulting in a total of 16 features \cite{koklu2020multiclass}. The outliers were removed for each class from the dataset using the Z-score method and the total size of the dataset was reduced to 12,909. The PCA dimension reduction technique was applied, reducing the total columns of the dataset to 10.

XGBoost and Support Vector Machine (SVM) classification models were created with nested cross-validation methods, and performance metrics were compared. Overall correct classification rates were determined as 94.00\% and 94.39\% for XGBoost and SVM respectively.

Dry bean is the most important and most produced pulse globally, and in Turkey, where it plays an important role in agriculture. The plant is sensitive to the effect of climatic changes, and resistance and/or tolerance to plant stress factors may be increased by breeding new seed cultivars and determining seed characteristics. The seed quality is an important factor in crop production, and automatic methods for grading and classification are required to ensure uniformity and efficiency. In Turkey, dry beans are divided into varieties based on their botanical characteristics, and the identification of bean varieties helps farmers to use seeds with basic standards for planting and marketing \cite{koklu2020multiclass}.

\section{Methodology}

\subsection{Exploratory data analysis (EDA)}
\subsubsection{Data Description}

\begin{figure}
\centering
\includegraphics[width=0.6\textwidth]{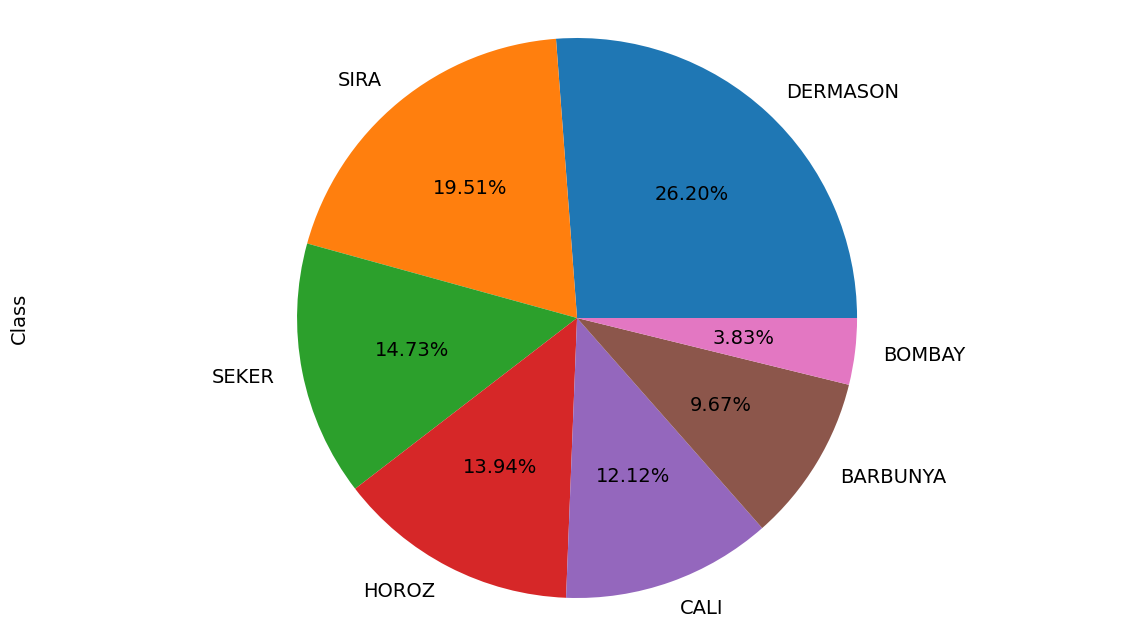}
\caption{\label{fig:pie_data_dist}
The data distribution among classes is depicted in the pie chart. It is evident that DERMASON represents the largest class, accounting for 26.20\% of the data, while BOMBAY constitutes the smallest class, representing only 3.83\% of the data.
}
\end{figure}

The original paper presents a comprehensive methodology for obtaining images of dry beans using a computer vision system. The  employed methodology employed for obtaining clear images of dry beans for multiclass classification involved the use of a computer vision system, image processing techniques, and segmentation using Otsu's method, resulting in a total of 16 features. The resulting dataset of 13,611 dry bean samples from seven basic varieties was captured using high-quality dimensional and shape features. To obtain the necessary samples, one kilogram of the seven basic dry bean varieties was procured from certified seed producers, in accordance with the standards set by the Turkish Standards Institute. Figure \ref{fig:pie_data_dist} illustrates the distribution of data among seven classes.

\begin{figure}
\centering
\includegraphics[width=0.6\textwidth]{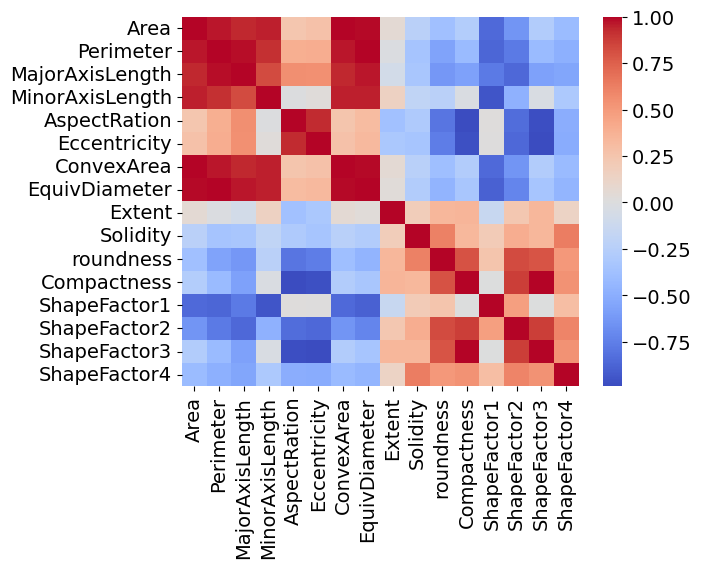}
\caption{\label{fig:data_corr}
Each cell shows the correlation between two variables. A correlation coefficient is a statistical measure that calculates the strength of the relationship between the relative movements of two variables. The values range between -1.0 and 1.0. A correlation of -1.0 shows a perfect negative correlation, while a correlation of 1.0 shows a perfect positive correlation. A correlation of 0.0 shows no linear relationship between the movement of the two variables.
}
\end{figure}

\subsubsection{Data Cleaning and Preprocessing}

\paragraph{Standardizing the data using StandardScaler}

In the subsequent data preprocessing step, we standardized the data using a Standard Scaler. Standardizing the features involves rescaling them to have a mean of 0 and a standard deviation of 1. This is also known as feature scaling or Z-score normalization. It is a crucial step before applying various machine learning algorithms, as many of these are sensitive to the scale of the input features. Standardizing the features so that they are centered around 0 with a standard deviation of 1 is particularly important when comparing measurements that have different units \cite{Raschka2015Python}.

The StandardScaler uses the formula:

\begin{center}
\[
x_{\text{scaled}} = \frac{x - \mu}{\sigma}
\]
\end{center}

In this formula, x represents the original value, $\mu$ denotes the mean of the data, $\sigma$ represents the standard deviation, and $x_{\text{scaled}}$ represents the scaled value of x. The formula subtracts the mean from the original value and then divides the result by the standard deviation to obtain the scaled value.

Standardization of a dataset is a common requirement for many machine learning estimators: they might behave badly if the individual features do not more or less look like standard normally distributed data (e.g., Gaussian with 0 mean and unit variance). For instance, many elements used in the objective function of a learning algorithm (such as the RBF kernel of Support Vector Machines or the L1 and L2 regularizers of linear models) assume that all features are centered around zero and have variance in the same order.

Standardization is essential for certain machine learning models like Principal Component Analysis (PCA), Logistic Regression, Support vector machines (SVMs), and more to behave optimally. After scaling, each feature in the dataset will have a zero mean and a standard deviation of one.

\paragraph{Using Z-Score to remove outliers}
The Z-Score tool measures the deviation of a specific data point from the dataset's mean in terms of standard deviations \cite{peck2011statistics}. The z-score, which may be positive or negative depending on whether the data point is above or below the mean, reflects the number of standard deviations that a data point is from the mean. Most large data sets have a Z-score between -3 and 3, meaning they lie within three standard deviations of the mean.

Z-Score was applied for each class of the dataset and removed rows that had Z-Score outside the most common -3 and +3 range. 702 rows were removed from the dataset during this process.

\paragraph{Reducing the dataset dimension using PCA}
A principal component analysis (PCA) is a sophisticated statistical technique that is primarily used in data science to reduce dimensionality. This approach entails transforming the original variables into a new set of orthogonal variables, referred to as the principal components, which are linear combinations of the original variables. These principal components encapsulate the maximum variance in the data, with the first principal component accounting for the most variance and each subsequent component accounting for lesser amounts \cite{pearson1901liii}. The advantages of PCA include its ability to mitigate the curse of dimensionality, reduce noise, and facilitate visualization of high-dimensional data. Furthermore, PCA can be particularly beneficial when used in combination with nested cross-validation for model selection and hyperparameter tuning. However, PCA also has limitations, including an assumption of linear relationships between variables, difficulty in interpreting transformed principal components, and potential loss of information due to the elimination of less important components. Furthermore, PCA is sensitive to the scaling of variables and may be unduly influenced by outliers, necessitating careful preprocessing of data \cite{jolliffe2002principal}.

The PCA was conducted on a scaled dataset, the data was previously scaled to ensure that all the features had the same range. The PCA model was created such that it retained 99.99\% of the variance from the original data. This was achieved by setting the n\_components parameter to 0.9999.

\begin{figure}
\centering
\includegraphics[width=0.5\textwidth]{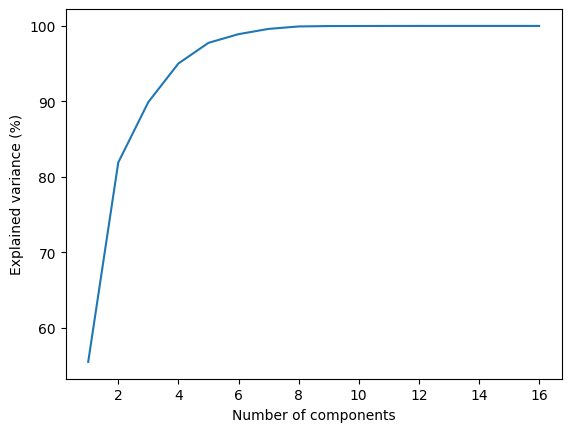}
\caption{\label{fig:PCA}This plot provides a visual representation of each component and the cumulative variance explained by each, starting with the first component. The x-axis of the plot represents the number of components, and the y-axis represents the cumulative explained variance in percentage. A total of 10 principal components were identified that cumulatively explained 99.99
}
\end{figure}

Following the PCA transformation, the cumulative variance explained by each component was calculated and plotted as a scree plot, as shown in Figure \ref{fig:PCA}.

In addition to identifying the number of components that captured 99.99\% of the variance, the analysis also aimed to identify the four most important features contributing to each component. The analysis showed that each component was primarily influenced by different subsets of features. The results can be seen in table \ref{tab:PCA_top4}.

\begin{table}[ht]
\centering
\begin{tabular}{|c|c|}
\hline
\textbf{Component} & \textbf{Top 4 Important Features}\\
\hline
Component 0 & MajorAxisLength, ShapeFactor2, Perimeter, EquivDiameter \\
\hline
Component 1 & MinorAxisLength, ShapeFactor1, AspectRation, Compactness \\
\hline
Component 2 & Solidity, ShapeFactor4, AspectRation, roundness \\
\hline
Component 3 & Extent, ShapeFactor4, Eccentricity, ShapeFactor3 \\
\hline
Component 4 & ShapeFactor4, roundness, Solidity, Extent \\
\hline
Component 5 & ShapeFactor1, ConvexArea, Area, Solidity \\
\hline
Component 6 & Eccentricity, roundness, ShapeFactor2, ShapeFactor1 \\
\hline
Component 7 & AspectRation, roundness, Eccentricity, Solidity \\
\hline
Component 8 & ShapeFactor2, MinorAxisLength, MajorAxisLength, ShapeFactor1 \\
\hline
Component 9 & MinorAxisLength, ConvexArea, Area, Perimeter \ \\ \hline
\end{tabular}
\caption{\label{tab:PCA_top4}Top 4 most important features in each component}
\end{table}

\paragraph{Label Encoding}
Label Encoding is a popular encoding technique for handling categorical variables. In this technique, each unique category value is assigned a unique integer.
In our dataset, we have a categorical variable with seven different classes, i.e., DERMASON, SIRA, SEKER, HOROZ, CALI, BARBUNYA, and BOMBAY. We applied label encoding to this categorical variable using the LabelEncoder class from the preprocessing module of the sklearn library.

\subsection{Models}
This research employed two distinct models. The first one, Support Vector Machine (SVM), was selected as the benchmark model due to its superior performance as highlighted in the original study. The second model was XGBoost, which was incorporated because it was absent in the initial paper and it typically delivers reliable results. 
\subsubsection{XGBoost}
Extreme Gradient Boosting (XGBoost) is a machine learning algorithm based on the gradient boosting framework. Gradient boosting is an ensemble technique where new models are added to correct the errors made by existing models. Models are added sequentially until no further improvements can be made \cite{chen2016xgboost}. The key advantage of XGBoost over other gradient boosting methods lies in its scalability, which drives its popularity among data scientists.

The XGBoost model was trained using a nested cross-validation technique, a procedure used to avoid overfitting when tuning hyperparameters. The inner loop is responsible for tuning the hyperparameters and selecting the best model, while the outer loop estimates the prediction error of the final chosen model.

The hyperparameters tuned during the nested cross-validation for the XGBoost model are shown in Table \ref{tab:model_parameters}.

\subsubsection{Support Vector Machine (SVM)}

Support Vector Machine (SVM) is a supervised machine learning algorithm that can be used for both classification and regression tasks. SVMs are particularly well suited for classification of complex, but small or medium-sized datasets \cite{cortes1995support}. The goal of SVM is to find a hyperplane in an N-dimensional space (N - the number of features) that distinctly classifies the data points.

As with XGBoost, the SVM model was also trained using a nested cross-validation technique. The same procedure and folds were followed as with the XGBoost model for consistency and comparison. The hyperparameters tuned during the nested cross-validation for the SVM model were kernel type, penalty parameter C, and gamma.

The kernel function is used to transform the input data into the required form. SVM has four types of kernels: linear, polynomial, radial basis function (RBF), and sigmoid.

The penalty parameter C of the error term determines the tradeoff between achieving a low training error and a low testing error, which is the bias-variance tradeoff. A high C aims at classifying all training examples correctly by giving the model freedom to select more samples as support vectors.

The gamma parameter defines how far the influence of a single training example reaches, with low values meaning 'far' and high values meaning 'close'. It can be seen as the inverse of the radius of influence of samples selected by the model as support vectors.

\subsection{Hyper-parameters Tuning}

\begin{table}[ht]
\centering
\label{tab:model_parameters}
\begin{tabular}{ll}
\hline
\textbf{Model} & \textbf{Hyperparameters} \\ \hline
XGBoost & \begin{tabular}[c]{@{}l@{}}'n\_estimators': [50, 100, 150],\\ 'learning\_rate': [0.1, 0.3],\\ 'colsample\_bytree': [0.3, 0.7, 1],\\ 'max\_depth': [10]\end{tabular} \\ \hline
SVM & \begin{tabular}[c]{@{}l@{}}'n\_components': [10],\\ 'C': [0.1, 1, 10],\\ 'kernel': ['linear', 'rbf'],\\ 'gamma': ['scale', 'auto']
\end{tabular} \\ \hline
\end{tabular}
\caption{Optimization Hyperparameters for XGBoost and SVM Models}
\end{table}

Nested cross-validation is used for training and optimizing parameters. Nested cross-validation is a method used in predictive modeling to estimate the performance of a model, particularly when tuning hyperparameters and estimating the prediction error of the final optimized model.
The procedure consists of two levels of cross-validation:

\begin{enumerate}
    \item \textbf{Outer cross-validation:} This is used to estimate the prediction error of the model. In each fold of the outer cross-validation, the data is split into a training set and a test set.
    \item \textbf{Inner cross-validation:} This is used to tune the model's hyperparameters. In each fold of the outer cross-validation, the training set is further divided into a training set and a validation set by the inner cross-validation. The model's hyperparameters are tuned based on the performance on the validation set.
\end{enumerate}

The nested cross-validation was performed using a 3-fold cross-validation for the inner loop and a 5-fold cross-validation for the outer loop. The grid search was used to tune the hyperparameters in the inner loop. The best model, i.e., the model with the highest accuracy, was selected and used for the outer loop. The average of the performance metric from the outer loop was used as the final performance metric for the model.

\begin{figure}
\centering
\includegraphics[width=0.5\textwidth]{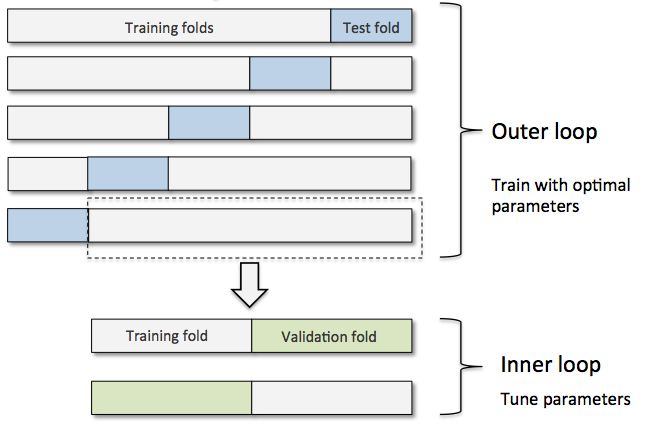}
\caption{\label{nested_cross_val}Illustration of the nested cross-validation process.}
\end{figure}

\section{Results}

The results obtained from nested k-fold cross-validation demonstrated that both the XGBoost and Support Vector Machine (SVM) models achieved high performance in classifying the seven different registered varieties of dry beans.

For the XGBoost model, the mean accuracy, F1 Score, and recall were all found to be 94\%. Five different confusion matrices were obtained from the nested cross-validation, giving an insight into the prediction distribution for each of the seven bean classes across different iterations. The overall best hyperparameters for the XGBoost model were a colsample\_bytree of 0.7, a learning\_rate of 0.1, and n\_estimators set to 100.

In comparison, the SVM model slightly outperformed the XGBoost model, yielding a mean accuracy, F1 Score, and recall of 94.39\%. Like XGBoost, five confusion matrices were obtained, indicating the SVM model's predictive power across different iterations of the cross-validation. The overall best hyperparameters for the SVM model were C set to 10, gamma as 'scale', and kernel as 'rbf'.

A close examination of the confusion matrices revealed that both models generally performed well in classifying the dry bean varieties. However, some classes were consistently misclassified, suggesting a level of difficulty in distinguishing between certain varieties. The models tended to misclassify varieties with similar features, indicating the need for more discriminative features or more complex models to handle these challenging cases.

\section{Discussion}
In the initial study \cite{koklu2020multiclass}, the authors did not provide any clear indication of the use of a validation set or methodology for hyperparameter optimization. Their methodology description only alludes to the application of K-fold cross-validation. This technique, according to \cite{arlot2010survey}, entails the random partitioning of the data set into equally sized subsets. One of these subsets is held out as a test set, and the model is trained on the remaining sets. This process is reiterated until all subsets have undergone testing. The results gleaned from this approach are subsequently generalized. For the study in question, the number of partitions or 'folds' (k) was set at 10.

Contrastingly, our study opted for the implementation of nested cross-validation, which we propose as a more precise and logical approach \cite{bates2021cross}. The nested cross-validation framework provides a more robust evaluation of the model's performance, especially in hyperparameter tuning, by eliminating the risk of overfitting to the validation set \cite{varma2006bias}. Consequently, we observed a slight deviation in our results as compared to the original paper, with a percentage difference ranging from 1 to 3 percent. Our study underscores the importance of carefully considering validation strategies and hyperparameter optimization techniques in the quest for reliable and reproducible research findings.
The two multiclass classification models, XGBoost and SVM, were employed in our study and demonstrated a high correct classification rate, 94.00\% and 94.39\% respectively. This indicates that both models were effective at distinguishing among the seven different registered varieties of dry beans. Despite the similarities in performance, the SVM model showed a slightly better overall classification rate. This could be attributed to the kernel trick in SVM, which allows for non-linear decision boundaries and can enhance the model's flexibility when dealing with complex data.

Our study also highlights the importance of pre-processing steps in classification tasks. The application of the Z-score method for outlier removal and PCA for dimensionality reduction not only cleaned the dataset but also improved computational efficiency without significant loss of information. These steps are critical in achieving reliable classification results, as the quality of the input data directly impacts the performance of the model.

The advancement in automated seed classification offers significant benefits for agricultural practices, especially in Turkey, where dry beans play an essential role. The ability to accurately identify and classify different bean varieties could enhance the seed selection process, leading to better crop yields, and greater resilience against climatic changes. This is of paramount importance given the recent climate change impacts on agriculture. Moreover, it helps in maintaining the standardization of seeds for planting and marketing, which is necessary for both domestic and international trade.

Nonetheless, it is important to acknowledge that the study was based on a specific dataset, which might limit the generalizability of the results. Future studies could incorporate more diverse datasets, including different bean varieties from different regions or countries, and consider other factors such as the impact of environmental conditions on the physical characteristics of the beans. Furthermore, the application of other machine learning algorithms and deep learning techniques, like Convolutional Neural Networks, could be explored to potentially improve classification accuracy.

In conclusion, our study successfully applies machine learning techniques to the classification of different dry bean varieties. The results underline the potential of these technologies in supporting agricultural practices and underline the necessity for careful validation and hyperparameter optimization strategies.

\bibliographystyle{alpha}
\bibliography{sample}

\end{document}